\documentclass{ecai} 

\usepackage{latexsym}
\usepackage{amssymb}
\usepackage{amsmath}
\usepackage{amsthm}
\usepackage{booktabs}
\usepackage{enumitem}
\usepackage{graphicx}
\usepackage{color}
\usepackage{array}
\usepackage{multirow}

\usepackage[T1]{fontenc}
\usepackage[utf8]{inputenc}
\usepackage{float}
\restylefloat{table}
\restylefloat{figure}
\usepackage{soul}

\newcommand{\BibTeX}{B\kern-.05em{\sc i\kern-.025em b}\kern-.08em\TeX}

\begin{document}


\begin{frontmatter}


\paperid{123} 


\title{Phrase-Level Adversarial Training for Mitigating  \\ Bias in Neural Network-based Automatic Essay Scoring}


 \author[A]{\fnms{Haddad }~\snm{Philip}}
 \author[B]{\fnms{Tsegaye Misikir }~\snm{Tashu} \thanks{Corresponding Author. Email: t.m.tashu@rug.nl}\footnote{Equal contribution.}
 \orcid{0000-0001-5793-9498}\footnotemark
 }

 \address[A]{Data Science and Engineering Department \\ Eötvös Lorand University, Budapest, Hungary}
 \address[B]{Artificial Intelligence Department\\
 Bernoulli Institute, University of Groningen, \\
 Nijenborgh 4, 9747 AG Groningen, Netherlands}


\begin{abstract}

Automatic Essay Scoring (AES) is widely used to evaluate candidates for educational purposes. However, due to the lack of representative data, most existing AES systems are not robust, and their scoring predictions are biased towards the most represented data samples. In this study, we propose a model-agnostic phrase-level method to generate an adversarial essay set to address the biases and robustness of AES models. Specifically, we construct an attack test set comprising samples from the original test set and adversarially generated samples using our proposed method. To evaluate the effectiveness of the attack strategy and data augmentation, we conducted a comprehensive analysis utilizing various neural network scoring models. Experimental results show that the proposed approach significantly improves AES model performance in the presence of adversarial examples and scenarios without such attacks.

\end{abstract}

\end{frontmatter}

\section{Introduction}

Automatic essay scoring systems (AES) are used to reduce the workload of examiners, improve the feedback cycle in the teaching-learning process, and save time and costs in grading \cite{Alikaniotis_2016,tashu2022deep}. Training usable AES models depends on the availability of balanced and representative data across different class labels (essay scores). Due to the unavailability of such representative data, most existing AES systems are not robust, and the scoring predictions are biased towards the most represented data samples. Therefore, increasing AES models' usability requires more robustly trained models with respect to out-of-domain data and underrepresented data samples \cite{acl-2020-association, Tashu2022,9031502}. Studies have shown that AES can be easily tricked by adding perturbations to the input, leading to incorrect predictions with high confidence \cite{szegedy2014intriguing}. These failures could hinder the safe deployment of these models in the real world and impact the trustworthiness of AES models.

Such perturbations can be added easily at different levels, with the most widely used being word-level perturbation. Several approaches have been proposed to address such challenges, improving resilience against these perturbations and reducing bias. Among these, data augmentation methods, which extend the training set to include adversarial examples, are widely used in the training process (fine-tuning) to improve the model's robustness. The problem of perturbations and the use of data augmentation have not been well addressed in the field of AES, especially at the phrase level. As the usage of AES systems is expanding from scoring essays for educational purposes to evaluating candidates \cite{singla2021aes} during interviews, minimizing biases and increasing the robustness of such models is relevant and creates trustworthiness towards the end users of the model. Therefore, in this study, we propose the use of a model-agnostic phrase-level method to generate an adversarial essay set to address the issue of biases and robustness of AES models. This study aims to train AES models on adversarially generated and augmented essays to build models that can score non-relevant generated essays properly.

\section{Related Work}

Automated Essay Scoring (AES) systems have become an important system in assessing student essays. With the adoption of advanced machine learning models including transformer-based architectures AES systems have become more effective. However, studies showed that neural network-based models are vulnerable to adversarial perturbations  \cite{singla2021aes,pham-etal-2021}.  These adversarial examples or attacks are created by minor changes in the input data which mislead the algorithms into making wrong decisions \cite{ebrahimi-etal-2018}. Some research has been done to address these vulnerabilities.

Gupta \cite{Gupta2023} explore the use of transformer models like BERT and RoBERTa for automated essay scoring (AES). It shows how data augmentation can help these models, especially with a wide range of essay topics. Their work showed that one model can be trained to grade essays across multiple subjects which is super helpful when there is limited data for certain topics and improves the model’s accuracy and generalizability. The work by Park et al. introduces EssayGAN \cite{Park2022}, an approach that uses generative adversarial networks (GANs) to generate synthetic essays. These essays are then scored and added to the existing datasets for AES systems. The results show that adding EssayGAN-generated essays to the training data improves the scoring of AES systems. Another work by \cite{jong2022} used data augmentation techniques like back-translation and score adjustment to improve AES models. The authors create a way to artificially add more essay-score pairs to the dataset and apply it to the Automated Student Assessment Prize dataset. The paper shows that data augmentation significantly improves the model’s performance by training various AES models on this augmented data. This includes better handling of language variations and robustness to overfitting.

The work by Tashu and Tomas\cite{Tashu2022} introduced a novel approach to enhancing the robustness of AES models against synonym-based adversarial attacks. Their focus is on synonym-based attacks, a specific type of adversarial attack that can be particularly challenging for AES models. This method involves generating new essays that are lexically similar to the original ones using keyword-based lexical substitution with BERT, and then using these adversarial samples to train the AES models.

Most of the existing works use semantic similarity to validate the perturbed examples. However, this method is unreliable in maintaining textual integrity as noted by Morris \cite{Morris_2020}.  Lei et al. \cite{lei2022phraselevel} have introduced methods to address the limitations of current approaches in preserving context-dependent content using label-preserving filters validated through human evaluation to enhance the fidelity of adversarially perturbed samples. This paper proposes a new model agnostic approach to generate phrase-level adversarial essays that extend and hybridise previous works \cite{lei2022phraselevel,Tashu2022}. By refining the training of AES models our approach aims to make the models robust to complex perturbations and to produce reliable scores across diverse and adversarially augmented datasets so that AES can be trusted and effective in educational and professional settings.

\section{MA-PLAG: Model Agnostic Phrase-Level Adversarial Generation}

\begin{figure*}[ht]
    \caption{The architecture of the proposed Model Agnostic Phrase-Level Adversarial Generation (MA-PLAG)}
    \centering
    \includegraphics[width=12cm, height=9cm]{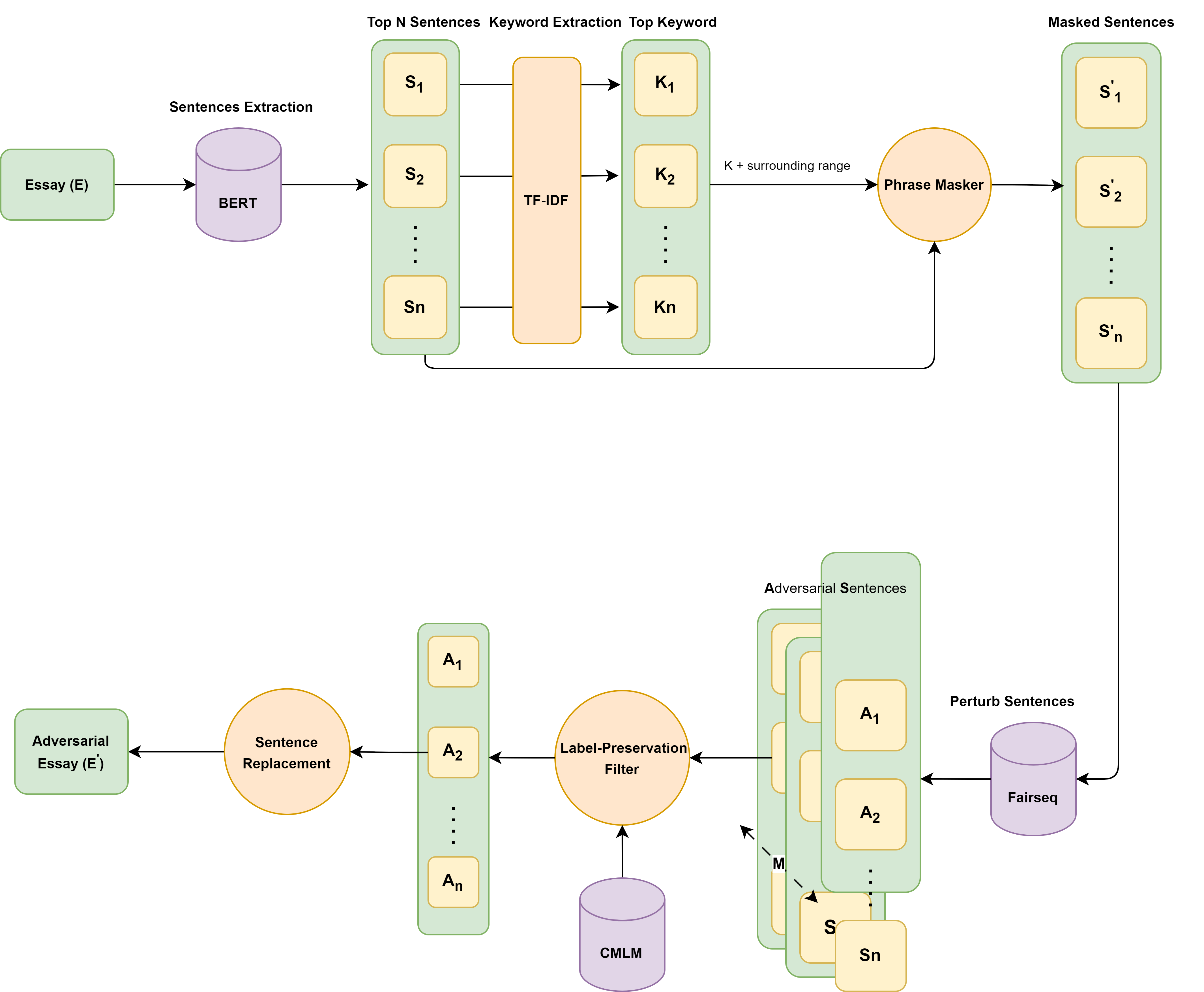}
    \label{fig:ma-plat-arch}
\end{figure*}

One of the most widely used strategies to minimize bias and to train a robust model when there is small and unbalanced data is to generate adversarial samples and augment the training data. In this study, a phrase-level adversarial data generation strategy is used. The overview of the proposed adversarial essay generation architecture is presented in Figure \ref{fig:ma-plat-arch}. The process is divided into several stages to generate label-preserved samples which have richer vocabulary required to train a more robust model.  Given an essay $E$ and to generate an adversarial essay $E'$ from it, the proposed model agnostic phrase-level adversarial generation approached works as follows: 

\begin{enumerate}
    \item The sentence extractor using the text summarizer in step \ref{sec:sent-extraction} will extract $N$ important sentences.
    \item The Phrase extracts in step \ref{sec:phrase-extraction} selects and extracts the relevant phrases from the sentences. 
    \item The blank-infilling model in step \ref{sec:blank-infilling} will generate $M$ perturbed phrases for each single sentence $s$, resulting in a set $\mathcal{S'}$ of perturbed sentences.
    \item  Label-preserving filter in step \ref{sec:label-pres-filter} will be applied on the set $\mathcal{S'}$, and only the relevant $s'$ will be kept.
    \item The final step is to replace each sentence $s$ in $E$ with the sentence $s'$, resulting in the adversarial essay $E'$.
\end{enumerate}

\subsection{Sentence Extraction}
\label{sec:sent-extraction}

Rather than randomly selecting sentences from a given essay, extracting the most important and content-bearing sentence from the given essay is beneficial. We used the extractive text summarization method to extract the most relevant sentence from the given essay. As the objective is to perturb the essay, it would be beneficial to reduce the text volume by condensing the information. We adopt the work by \cite{miller2019leveraging} which uses BERT and K-Means for extractive text summarization.

\subsection{Phrase Extraction}
\label{sec:phrase-extraction}

For each sentence selected by the summarization algorithm, we extract the most significant keyword using the term frequency/inverse document frequency (TF-IDF) approach. These extracted keywords serve as the central element around which the phrase will be constructed. Considering the sentence length, typically ranging from 20 to 25 words, we adopt a phrase definition strategy that incorporates the TF-IDF keyword and the surrounding words within a specific range of length denoted by the parameter $N$. The determination of $N$ is treated as a hyperparameter and can be adjusted according to specific requirements. 
Our phrase extraction approach gives MA-PLAG the flexibility to address a wide range of problem domains. Let $s$ be a sentence in the essay $E$, and $w$ be a word in $s$. The Keyword Extraction can be defined as follows:

    \begin{equation}
        w^* = \arg\max_{w \in s} \text{TF-IDF}(w, s, E)
    \end{equation}
    where $w^*$ is the most significant keyword in sentence $s$ based on TF-IDF values. The Phrase Construction can be defined as follows:
    \begin{equation}
        P(w^*, s, N) = s[i^* - N : i^* + N + 1]
    \end{equation}
    where $i^*$ is the index of $w^*$ in $s$, and $N$ is the hyperparameter defining the number of words before and after $w^*$ to include in the phrase $P$. The values of $N$ and $i^*$ are adjusted to ensure they fall within the valid indices of $s$.

\subsection{Sentence Perturbation}
\label{sec:blank-infilling}

\begin{figure}[H]
    \centering
    \includegraphics[width=\linewidth]{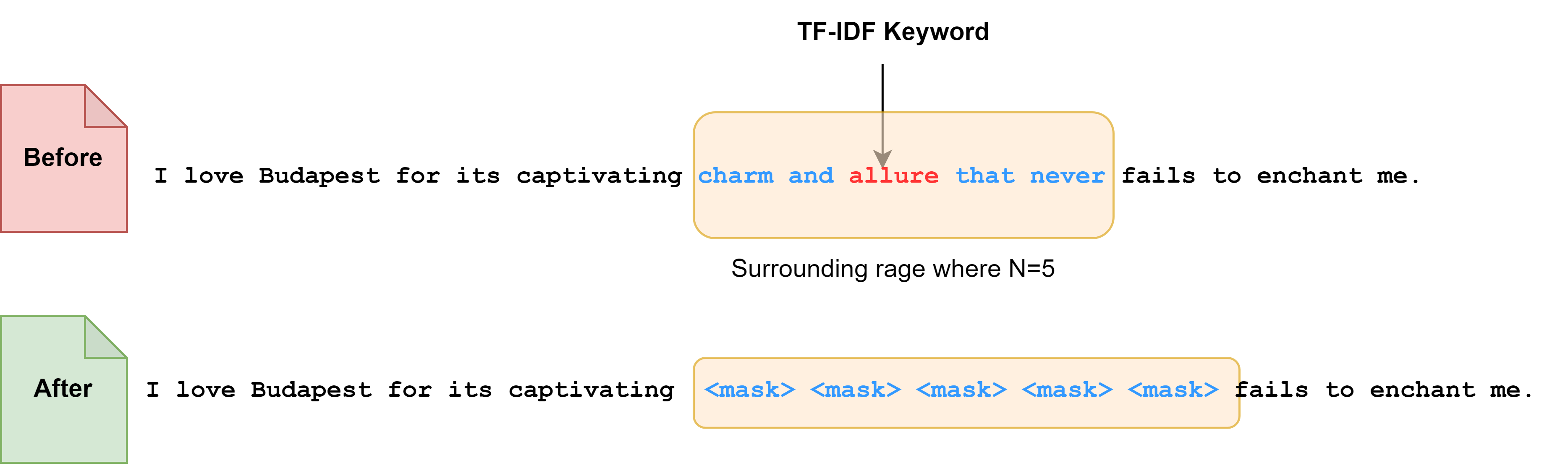}
    \caption{Sentence; before and after masking the phrase}
    \label{fig:mask-sentence}
\end{figure}

To perturbate each sentence,  a method called blank-infilling on each target phrase was adopted \cite{lei2022phraselevel}. Firstly, we mask the phrase with a suitable token depending on the blank-infilling procedure, as shown in Figure \ref{fig:mask-sentence}.

Subsequently, a pre-trained language model designed for blank-infilling tasks can be employed. Considering the surrounding context, this model takes the masked sentence as input and generates a phrase of $M$ tokens to fill the masks. In our case, we used Fairseq \footnote{\texttt{https://github.com/facebookresearch/fairseq}} as an implementation of the blank-infilling model BART \cite{lewis2019bart}. The blank-infilling model will generate perturbed sentences by replacing the masked phrase with "different" new phrases, and the maximum length of the new phrases is not greater than the length of the original phrase plus a threshold. Let \( S \) be a sentence from a document, and \( S' \) be the version of \( S \) where a phrase \( P \) is masked. Let \( B \) represent a blank-infilling model. The sentence perturbation process can be defined as follows:

\begin{equation}
S'' = 
\begin{cases} 
    \text{replace}(S', \text{mask}, B(S')) & \text{if } |B(S')| \leq |P| + \theta \\
    S' & \text{otherwise}
\end{cases}
\end{equation}
  where \( S'' \) is the final output sentence, \( |P| \) is the length of the original masked phrase, \( |Q| \) is the length of the generated phrase, and \( \theta \) is a threshold for the maximum permissible length of \( Q \) relative to \( P \).

\subsection{Label-Preserving Filter}
\label{sec:label-pres-filter}

To ensure that the true label is preserved, we adopted the same strategy called class-conditioned masked language models (CMLMs) \cite{lei2022phraselevel} to choose the most suitable perturbed sentence. In our case, we fine-tuned the RoBERTa Model \cite{liu2019roberta}, which is an extension of BERT. Since the AES task is being handled as a classification task, the true label, or alternatively the class-related characteristic, must be retained. To do this, a filter is applied directly to the perturbation using the likelihood supplied by the class-conditioned masked language models (CMLMs). Let's consider a sentence $s$ of length $n$, a phrase $p$, and $q$ as the perturbed phrase of length $m$ filled by the model. The perturbed sentence will be \(\tilde{s}_q(i,j) = s_1, \ldots, s_{i-1}, q_1, \ldots, q_m, s_{j+1}, \ldots, s_n\), where $(i,j)$ represent the starting and ending indices of phrase $p$ within the original sentence $s$, and \(\tilde{s}_q(i,j)\) represents the version of sentence $s$ where the phrase at indices $(i,j)$ has been replaced with the perturbed phrase $q$. The equation for the class-conditioned likelihood of the adversarially perturbed phrase $q$ in the sentence $s$ can be calculated as:

\begin{equation}
    L(s, q, y)=\prod_{k=1}^m P_{\mathrm{CMLM}}\left(q_k \mid \tilde{s}_{q \backslash q_k} ; \Theta_y\right).
\end{equation}

Where $\tilde{s}_{q \backslash q_k}$ is the sentence $\tilde{s}_q(i,j)$ with the token $q_k$ masked, and $P_\mathrm{CMLM}$ is the likelihood of $q_k$ given $\tilde{s}_{q \backslash q_k}$, which is produced by a class-conditioned masked language model $\Theta_y$ conditioned on class $y$.

To avoid label changing from the true label, the phrase perturbations should get a higher likelihood on the true class’s distribution and a lower likelihood on the others. This can be calculated as follows, where $\mathcal{Y}$ denotes the set of classes:

\begin{equation}
    R(s, q, y)=L(s, q, y) / \max _{\tilde{y} \in \mathcal{Y}, \tilde{y} \neq y} L(s, q, \tilde{y})
\end{equation}

For better label preservation, the chosen phrase perturbations are required to have a likelihood ratio larger than a certain threshold $\delta$.

\section{Scoring Models}
\label{sec:scoring-models}

To thoroughly evaluate the efficiency of our proposed adversary generation methodology, both as an attack strategy and a data augmentation technique, we conducted a comprehensive analysis utilizing a variety of neural network scoring models. The models have an Embedding Layer, CNN layer, RNN layer and Desne layer. We used GloVe and BERT, two popular pre-trained language models in the embedding layer to transform the input essay into dense numerical representations, capturing semantic and contextual information.
CNN was used to extract local features from the embedded representations using convolutional filters. It efficiently captured patterns and relevant information within the text. Bidirectional Long Short-Term Memory (Bi-LSTM) and Bidirectional Gated Recurrent Unit (Bi-GRU) were also used to capture sequential dependencies and contextual information over longer text sequences. The bi-directional approach allowed the models to effectively incorporate information from both past and future contexts, enhancing their understanding of overall text semantics. To maintain consistency across all models, we used a uniform Dense layer architecture. This layer served as the final stage for score prediction, transforming the extracted features into a single scalar value. 

We utilized the mean squared error (MSE) as the loss function to measure the disparity between the predicted and target scores. MSE was chosen for its property of penalizing larger errors more severely, thereby emphasizing accurate score prediction. We used the RMSProp optimization algorithm, which adaptively adjusts the learning rate based on the gradient history. This approach facilitated efficient convergence and improved the overall training performance of the models. By employing these well-defined architectures and methodologies, we constructed four distinct neural network models. Each model offered a unique combination of embedding techniques, convolutional and recurrent layers, and a shared dense layer, enabling a robust evaluation of our adversary generation methodology.
\section{Experiments}

\subsection{Datasets and Evaluation Strategy}

The experiments were performed on the dataset provided within the Automated Essay Scoring competition on the Kaggle \footnote{\texttt{https://www.kaggle.com/c/asap-aes}} website. The datasets contain student essays for eight different prompts (essay discussion questions). Four datasets included essays of traditional writing genres such as persuasive, expository, or narrative. The other four datasets were source-based, i.e., the students had to discuss questions referring to a previously read source document. At least two human expert graders score each training set. The authors of the datasets already divided them into fixed training and test sets.

\subsection{Evaluation Metric}

Quadratic Weighted Kappa (QWK or $\kappa$) metric is a widely used evaluation measure in the field of automatic essay scoring (AES) since it provides a quantitative assessment of the agreement between human-graded scores and the scores predicted by an AES system. QWK is designed to handle the ordinal nature of the essay scores where essays are typically assigned scores on a discrete scale, such as a numerical scale or a set of predefined categories.

\subsubsection{Training and Evaluation Methodology}

The dataset was split into three sets training, validation, and testing where 60\% was used for training, 20\% for evaluation, and 20\% for testing. Subsequently, the models discussed in Section \ref{sec:scoring-models} were trained on the training set and individually fine-tuned by adjusting the layers, learning rate and the number of epochs. The best model with the highest Quadratic Weighted Kappa (QWK) value on the validation set was selected and finally evaluated using the test set. The variation in performance of the model across different prompts can be observed from the results presented. 

\subsection{Applying Attack Strategy}

To assess the effectiveness of the proposed phrase-level attack strategy and augmentation, we formulated an attack test set comprising a combination of samples extracted from the original test set and those generated adversarially using our method. The primary objective of this test set was to evaluate the degree to which these samples were able to deceive the pre-trained models. To enhance the robustness of our experimental evaluation, we incorporated two key aspects in the construction of this attack set:

\begin{itemize}
    \item Firstly, at the generation level, we applied varying ratios to mask individual sentences before inputting them into the blank-infilling model. For instance, given a sentence $S$ with a length of 20 words and a generation ratio of 30\%, this would mask six words from sentence $S$ before supplying it to the blank-infilling model. The resulting sentence would contain 70\% of the original, and the rest would be generated words.
    \item Secondly, we explored different attack sample sizes at the combination level while building the attack set. Suppose we have a data set with 100 samples, and the attack size ratio is set at 50\%. In this scenario, we will generate 50 attack/adversarial samples to yield 150 samples within the attack set. In this phase, we considered the class imbalance issue so that more samples would be generated for underrepresented classes.
\end{itemize}

\subsection{Data Augmentation}

Given the objective of constructing robust models capable of withstanding sample perturbations and bias, we propose the adoption of phrase-level data augmentation as a strategy to address the problem. Specifically, we augment the adversarial examples generated with the original dataset in a manner consistent with the methodology employed during the creation of the attack set. Subsequently, we subject our AES models to retraining and evaluate their performance using the designated test set.

\section{Experiment Results and Discussion}

In this experiment, we mainly focused on the following masking and generation ratios: 30\% and 40\%, and with the adversarial attack generation and augmentation size of 50\% and 75\%. "No Attack" refers to the models being trained and tested on the original data, "With Augmentation" refers to the models being trained and tested on the augmented data, and "With Attack" means that the models were trained on the original data and tested on the adversarial attack data.

\begin{table}[h]
\caption{Results for BERT model with a generation ratio of 30\% and attack size of 50\%.}
\centering
  \begin{tabular}{llll}
  \toprule
    \textbf{Prompt} & \textbf{No Attack} & \textbf{With Attack} & \textbf{With Augmentation} \\
    2 & 0.613 & 0.346 & 0.610 \\
    3 & 0.762 & 0.707 & 0.827 \\
    4 & 0.694 & 0.683 & 0.808 \\
    5 & 0.856 & 0.783 & 0.907 \\
   \bottomrule
  \end{tabular}
\end{table}

\begin{table*}[h]
  \centering
  \caption{Results for Bi-LSTM \& Bi-GRU models with a generation ratio of 30\% and attack size of 50\%}
  \begin{tabular}{ccccccc}
    \toprule
    \multirow{2}{*}{\textbf{Prompt}} & \multicolumn{2}{c}{\textbf{No Attack}} & \multicolumn{2}{c}{\textbf{With Attack}} & \multicolumn{2}{c}{\textbf{With Augmentation}} \\
    & \textbf{Bi-LSTM} & \textbf{Bi-GRU} & \textbf{Bi-LSTM} & \textbf{Bi-GRU} & \textbf{Bi-LSTM} & \textbf{Bi-GRU} \\
    2 & 0.695 & 0.696 & 0.590 & 0.587 & 0.764 & 0.801 \\
    3 & 0.756 & 0.695 & 0.607 & 0.572 & 0.769 & 0.760 \\
    4 & 0.715 & 0.757 & 0.614 & 0.605 & 0.798 & 0.790 \\
    5 & 0.813 & 0.812 & 0.701 & 0.708 & 0.849 & 0.817 \\
    \bottomrule
  \end{tabular}
\end{table*}

\begin{table*}[h]
  \centering
  \caption{Results for Bi-LSTM \& Bi-GRU models with a generation ratio of 30\% and attack size of 75\%}
  \begin{tabular}{c c cc c c c}
    \toprule
    \multirow{2}{*}{\textbf{Prompt}} & \multicolumn{2}{c}{\textbf{No Attack}} & \multicolumn{2}{c}{\textbf{With Attack}} & \multicolumn{2}{c}{\textbf{With Augmentation}} \\
    & \textbf{Bi-LSTM} & \textbf{Bi-GRU} & \textbf{Bi-LSTM} & \textbf{Bi-GRU} & \textbf{Bi-LSTM} & \textbf{Bi-GRU} \\
    
    2 & 0.688 & 0.732 & 0.542 & 0.550 & 0.756 & 0.811 \\
    3 & 0.689 & 0.711 & 0.589 & 0.611 & 0.793 & 0.807 \\
    4 & 0.744 & 0.689 & 0.605 & 0.568 & 0.738 & 0.762 \\
    5 & 0.830 & 0.828 & 0.711 & 0.717 & 0.843 & 0.874 \\
    \toprule
  \end{tabular}
\end{table*}

\begin{table}[h]
  \centering
  \caption{Results for BERT model with a generation ratio of 30\% and attack size of 75\%}
  \begin{tabular}{cccc}
    \toprule
    \textbf{Prompt} & \textbf{No Attack} & \textbf{With Attack} & \textbf{With Augmentation} \\
    
    2 & 0.615 & 0.354 & 0.609 \\
    3 & 0.759 & 0.700 & 0.797 \\
    4 & 0.695 & 0.692 & 0.803 \\
    5 & 0.837 & 0.783 & 0.897 \\
    \bottomrule
  \end{tabular}
\end{table}

\begin{table}[h]
  \centering
  \caption{Results for BERT model with a generation ratio of 40\% and attack size of 50\%}
  \begin{tabular}{cccc}
    \toprule
    \textbf{Prompt} & \textbf{No Attack} & \textbf{With Attack} & \textbf{With Augmentation} \\
    
    2 & 0.611 & 0.302 & 0.617 \\
    3 & 0.757 & 0.628 & 0.828 \\
    4 & 0.695 & 0.616 & 0.756 \\
    5 & 0.837 & 0.722 & 0.882 \\
    \bottomrule
  \end{tabular}
\end{table}

\begin{table*}[h]
  \centering
  \caption{Results for Bi-LSTM \& Bi-GRU models with a generation ratio of 40\% and attack size of 50\%}
  \begin{tabular}{ccccccc}
    \toprule
    \multirow{2}{*}{\textbf{Prompt}} & \multicolumn{2}{c}{\textbf{No Attack}} & \multicolumn{2}{c}{\textbf{With Attack}} & \multicolumn{2}{c}{\textbf{With Augmentation}} \\
    & \textbf{Bi-LSTM} & \textbf{Bi-GRU} & \textbf{Bi-LSTM} & \textbf{Bi-GRU} & \textbf{Bi-LSTM} & \textbf{Bi-GRU} \\
    
    2 & 0.688 & 0.732 & 0.595 & 0.593 & 0.742 & 0.744 \\
    3 & 0.689 & 0.711 & 0.622 & 0.654 & 0.760 & 0.737 \\
    4 & 0.744 & 0.689 & 0.617 & 0.709 & 0.714 & 0.767 \\
    5 & 0.830 & 0.828 & 0.727 & 0.609 & 0.807 & 0.638 \\
    \bottomrule
  \end{tabular}
\end{table*}

The experimental results showed a notable decrease in model performance when adversarial attacks were used as a test set to evaluate models trained on the original trained data, indicating the vulnerability of the models to such attacks. All the models trained with original data and evaluated using the adversarial test set (referred to as the attack set) in all experiments showed a significant decrease in performance. This serves as evidence that our attacks can easily deceive these models. The adversaries' attacks had a considerable impact on the classification accuracy of the models. Specifically, for the Bi-LSTM-based model, the accuracy deteriorated by 0.105, 0.149, 0.101, and 0.112 for test sets 2, 3, 4, and 5, respectively. Similarly, the Bi-GRU-based model experienced accuracy deterioration of 0.109, 0.123, 0.152, and 0.104 for the corresponding test sets. The accuracy of the BERT model exhibited decreases of 0.267, 0.055, 0.011, and 0.073 for essay sets 2, 3, 4, and 5, respectively. 

Consequently, the classification accuracy declined across all four datasets utilized in the experiment. Notably, the most substantial deterioration occurred in the case of the BERT model, with a decrease of 0.267 in accuracy for essay set 2. This could be attributed to the fact that our BERT model\footnote{For technical resource concerns, we used DistilBERT} has a maximum sequence length of 512, and around 25\% of samples in set 2 are longer than 512 words per sequence.

The experiments indicate that the inclusion of adversarial examples into the original dataset, along with their utilization as a defence strategy, can enhance the performance of AES models. Specifically, when assessing the impact on classification accuracy, the Bi-LSTM-based model exhibited improvements of 0.174, 0.162, 0.083, and 0.066 for essays 2 through 5, respectively, after augmenting the dataset with adversarial examples. Similarly, the Bi-GRU-based model demonstrated enhanced classification accuracy of 0.194, 0.236, 0.173, and 0.159 for essays 2 through 5, respectively, compared to models that did not undergo adversarial attacks.
By training the models with augmented examples generated using our proposed method, their overall performance on the combined test set showed average improvements of 0.163, 0.167, and 0.148 for Bi-LSTM, Bi-GRU, and BERT-based models, respectively.

\subsection{Ablation Study}

An ablation study was conducted to observe which component of the attack affects the performance of the model. Tables 1, 2, 3, 4, 5 and 6 depict the classification accuracy of the attack on Bi-LSTM, Bi-GRU and BERT models while varying the generation ratio and the attack size.  In the first case, we varied the generation ratio from 30\% to 40\% while maintaining the attack size at 50\% and compared the classification accuracy of the attacks. In the second case, we varied the attack size while maintaining the generation ratio at 30\% and compared the classification accuracy of the attacks. The results of the ablation study show that both generation ratio and attack size are important factors.

When the generation ratio was increased from \textbf{30\% to 40\%}, the accuracy of both Bi-LSTM and Bi-GRU models slightly dropped. On the other hand, the performance of trained with augmentation remained almost unchanged implying that the augmentation process is insensitive to variation in generation ratio. For BERT, the performance of the model with a higher generation ratio of \textbf{40\%} is more pronounced for the “With Attack” case, especially for Prompt 2, where the accuracy dropped from \textbf{0.346 to 0.302}. This implies that BERT is more sensitive to variation in generation ratio than Bi-LSTM and Bi-GRU.

When the attack size was increased from \textbf{50\% to 75\%}, the Bi-LSTM and Bi-GRU models were noticeably affected. The accuracy of the models in the “With Attack” case dropped more for higher attack sizes thus showing that higher attack sizes yield more effective adversarial attacks. BERT also recorded decreased accuracy for higher attack sizes but the effect was less compared to the effect of changing the generation ratio. This shows that BERT is sensitive to generation ratio but relatively more stable to increase in attack size.

Generally, this shows that both generation ratio and attack size should be carefully chosen in devising adversarial attacks and defence mechanisms. The generation ratio seems to have more effect on the BERT model than the attack size. On the other hand, the attack size has more effect on the Bi-LSTM and Bi-GRU models than the generation ratio. These observations can thus be employed to devise more effective AES models by balancing between the generation of adversarial examples and their attack strengths.

\section{Conclusion}

This study introduced a unified solution to counteract biases and improve the robustness of Automatic Essay Scoring (AES) models via a new phrase-level adversarial training method. By leveraging a model-agnostic approach to produce adversarial essay sets and using data augmentation, we showed how augmenting the training data can greatly improve the performance and robustness of different neural network-based AES models. Experiment results showed how AES models are sensitive to adversarial attacks and the accuracy significantly dropped when models trained on original datasets were tested on adversarially generated test sets. However, models retrained on the augmented datasets that included adversarial examples showed restored and even improved performance, as ranked scores produced by them achieved higher Quadratic Weighted Kappa (QWK).

The paper also discussed the need for a trade-off between generation ratio and attack size for adversarial training. It showed that higher generation ratios and attack sizes (in general) decreased the accuracy of models, but the effective data augmentation due to adversarial training mitigated the reduction, resulting in higher robustness. Finally, we can say that our proposed phrase-level adversarial training method using data augmentation is a viable solution to improve the robustness of AES systems against biases and adversarial attacks. It guarantees AES models offer more accurate and unbiased essay scores and thus enhance their reliability and validity for real-world applications in educational institutions. As a by-product, this research can be extended to other natural language processing tasks to verify the effectiveness of this methodology in different domains.

\bibliography{main}

\end{document}